\pdfoutput=1

\documentclass[11pt]{article}

\usepackage[final]{acl}

\usepackage{times}
\usepackage{latexsym}
\usepackage{booktabs}
\usepackage{arydshln, caption}

\usepackage{multirow}
\usepackage{arydshln}

\usepackage{amssymb}
\usepackage{amsmath}
\usepackage{pifont}
\newcommand{\cmark}{\ding{51}}%
\newcommand{\xmark}{\ding{55}}%

\usepackage[T1]{fontenc}

\usepackage[utf8]{inputenc}

\usepackage{microtype}

\usepackage{inconsolata}
\usepackage{hyperref}

\usepackage{graphicx}
\usepackage{floatrow}
\newfloatcommand{capbtabbox}{table}[][\FBwidth]

\usepackage{tcolorbox} 
\usepackage{todonotes}

\definecolor{ocr}{HTML}{00C8FF}
\definecolor{ocr}{HTML}{009900}
\definecolor{jk}{rgb}{0.6, 0.2, 1.0}
\definecolor{jack}{rgb}{1.0, 0.6, 0.4}
\definecolor{owencolor}{rgb}{0.5, 0.2, 0.8}

\title{Evaluating LLMs with Multiple Problems at once}

\author{Zhengxiang Wang \and Jordan Kodner \and Owen Rambow\\
Department of Linguistics \& Institute for Advanced Computational Science\\
    Stony Brook University, Stony Brook, NY, USA \\
\texttt{\{first.last\}@stonybrook.edu} }

\begin{document}
\maketitle

\begin{abstract}

This paper shows the benefits and fruitfulness of evaluating LLMs with multiple problems at once, a paradigm we call \textit{multi-problem evaluation} (MPE). Unlike conventional \textit{single-problem evaluation}, where a prompt presents a single problem and expects one specific answer, MPE places multiple problems together in a single prompt and assesses how well an LLM answers all these problems in a single output. Leveraging 6 classification and 12 reasoning benchmarks that already exist, we introduce a new benchmark called ZeMPE (\textbf{Ze}ro-shot \textbf{M}ulti-Problem \textbf{E}valuation), comprising 53,100 zero-shot multi-problem prompts. We experiment with a total of 13 LLMs from 5 model families on ZeMPE to present a comprehensive and systematic MPE. Our results show that LLMs are capable of handling multiple problems from a single data source as well as handling them separately, but there are conditions this multiple problem handling capability falls short. In addition, we perform in-depth further analyses and explore model-level factors that may enable multiple problem handling capabilities in LLMs. We release our corpus and code\footnote{\url{https://github.com/jaaack-wang/multi-problem-eval-llm}} to facilitate future research.

\end{abstract}


\section{Introduction}

Thanks to the advances in both GPU hardware and algorithms \citep[][\textit{inter alia}]{dai-etal-2019-transformer, beltagy2020longformer, flashattn, ding2024longrope, chen2024longlora}, large language models (LLMs) have been developed with increasingly larger context windows (e.g., 8K, 128K, 2M). To leverage the extended context windows, recent studies \cite{cheng-etal-2023-batch, lin2024batchpromptaccomplish, son-etal-2024-multi-task} have proposed various prompting strategies that place multiple problems in a single prompt, which we collectively call \textit{multi-problem prompting} (MPP). The basic idea of MPP is to place multiple problems after a shared context $C$ (e.g., task instruction and/or exemplars) to avoid repeating $C$ for each problem as in standard \textit{single-problem prompting} (SPP), which improves input token utilization and reduces LLM inference costs per problem.

In this study, we evaluate a wide range of LLMs with multiple problems at once through MPP, a paradigm we call \textit{multi-problem evaluation} (MPE) \cite{wang-etal-2025-exploring}.\footnote{While MPE is achieved through MPP, MPP can be used for purposes other than evaluation, e.g., knowledge retrieval, question answering, and other use cases. It it thus necessary to distinguish MPP from MPE and SPP from SPE.} While the main goal of MPP is to improve the cost-efficiency of LLM inference, we view MPE primarily as a valuable evaluation paradigm for probing LLM capabilities, rather than merely a cost-saving engineering strategy. Unlike conventional \textit{single-problem evaluation} that assesses an LLM's ability to answer a single problem through SPP, MPE assesses an LLM's ability to concurrently handle multiple problems at once or in a single output. Understanding the multiple problem handling capabilities of LLMs is an important research question because it gives us a foundational insight into how LLMs operate over multi-problem inputs that can be sufficiently long and use information from individual problems contained within each multi-problem input.

To enable a comprehensive and systematic MPE, we introduce ZeMPE (\textbf{Ze}ro-shot \textbf{M}ulti-Problem \textbf{E}valuation), a new benchmark comprising 53,1000 zero-shot multi-problem prompts. ZeMPE is synthetically generated by leveraging 6 classification and 12 reasoning benchmarks that already exist and are widely used. Moreover, ZeMPE includes various types of evaluation tasks to allow for deep and nuanced analyses, taking into account how multiple problems are presented in the prompt and whether these problems are sampled from the same data source or not. We do not mix classification and reasoning problems together due to the different natures of these two types of problems and to not make our experiments confounding.

Our main contributions are as follows:

\begin{itemize}

    \item We show that LLMs are capable of handling multiple classification or reasoning problems from a single data source as well as handling them separately zero-shot. We present multiple pieces of evidence, in addition to direct performance comparisons to validate this. 

    \item We demonstrate that just like few-shot MPP, zero-shot MPP can be highly cost-efficient. 

    \item We identify two general conditions under which LLMs perform significantly worse than expected when presented with multiple problems and explore the roles of several model-level factors that may enable their multiple problem handling capabilities. 

    \item We release a new MPE benchmark called ZeMPE to facilitate future MPE studies.

\end{itemize}

\section{Related Work\label{sec:literatureReview}}

We note that current LLM evaluation has predominantly focused on LLM's performance on single-problem prompts. Each of such prompts presents a single problem and expects one specific answer to that problem, which may \textit{implicitly} require multi-hop reasoning or multi-step task solving. 

Recently, \citet{cheng-etal-2023-batch} propose few-shot MPP named batch prompting that prompts LLMs with problems batched together from single sources following a few batches of equally sized exemplars. They find that few-shot MPP greatly increases LLM inference efficiency while retaining downstream performance with a small batch size (e.g., <6). To ensure that batch prompting works with large batch sizes, \citet{lin2024batchpromptaccomplish} introduce a sampling optimization method that takes a majority vote over repeated permutations of batch samples.

Instead of solving multiple \textit{separate} problems, \citet{son-etal-2024-multi-task} prompt LLMs with exemplars to solve multiple \textit{related} tasks based on a shared problem setup by placing an \textit{explicit} instruction for each task. They find that instructing LLMs to solve all the tasks at once outperforms solving the individual tasks one by one or in a batch. 


In addition to these few-shot studies, \citet{laskar-etal-2023-systematic} shows that instruction-tuned GPTs can handle 5 short questions sampled from two open-domain QA benchmarks at once zero-shot, but the base GPT models can barely perform the task. To the best of our knowledge, \citet{wang-etal-2025-exploring} present the first systematic evaluation of LLMs’ zero-shot ability to tackle multiple homogeneous classification problems drawn from six standard benchmarks. They show that, while LLMs can usually solve several such classifications in a single prompt with accuracy comparable to handling them one-by-one, their performance deteriorates sharply when the prompt instead asks them to return the indices of texts belonging to each class--a shortfall that remains consistent across models, prompting conditions, and experimental settings.

Building on top of \citet{wang-etal-2025-exploring}, this study examines a total of 13 LLMs on 18 existing benchmarks, including 12 reasoning benchmarks that are not part of \citet{wang-etal-2025-exploring}'s evaluation. Besides from reaffirming \citet{wang-etal-2025-exploring}'s finding that LLMs are capable of handling multiple problems from a single data source as well as handling them separately, we perform in-depth
further analyses to both validate such capabilities and expose their limitations. Moreover, we explore model-level factors that may enable LLM's strong multiple problem handling capabilities. 


\section{Multi-Problem Evaluation \label{sec:mpe}}

This section compares single-problem evaluation (SPE) and multi-problem evaluation (MPE) and introduces ZeMPE, a new MPE benchmark.

\subsection{SPE vs. MPE}

SPE assesses an LLM's ability to solve a type of problem by prompting the LLM with such a problem one at a time. In contrast, MPE places multiple problems together that can be of a same or different types and evaluates how well an LLM handles them. A simple example of a multi-problem task would bundle multiple classification or QA problems together and ask LLMs to solve them sequentially.

\subsection{Benefits of MPE}

MPE has at least three advantages over SPE.

\paragraph{Lesser Data Contamination Concerns} First, it is less likely for LLMs to encounter exact multi-problem prompts during pre-training because of the combinatory nature of constructing prompts from multiple problems. This helps mitigate a growing data contamination concern in modern large-scale pre-training \citep{jacovi-etal-2023-stop, sainz-etal-2023-nlp}. 

\paragraph{Improved Controllability and Interpretability of Evaluation} Second, since we can manipulate \textit{what kind of problems} and \textit{how many problems} to include, we know exactly which problem an LLM gets wrong or right across positions in the prompts. This enables us to construct a well controlled and easily interpretable evaluation. 

\paragraph{High Feasibility and Adaptability} Third, our study demonstrates that leveraging the rich existing benchmarks to create a new multi-problem task is \textit{cheap}, \textit{easy to implement}, and \textit{highly adaptable}. The most laborious component is the prompt design, which, once done, can easily be applied to a set of benchmarks with minimal adaptation.

    

     
     
    
    

\begin{table*}[htb]
    \centering
    \scriptsize
    \begin{tabular}{p{1.75cm}p{2.75cm}p{4.75cm}p{1.25cm}p{2.75cm}}
    \toprule
    Problem Type  & Input/Output Format & Benchmark & \# Examples & Objective \\ \midrule
    
    Classification & Single-text input & SST-2 \citep{sst-2} & 1,821 & Sentiment analysis \\ 

     &  & CoLA \cite{cola} & 1,043 & Grammatical acceptability \\ 
     &  & AGNews \citep{AGNews} & 1,000 & Topic classification \\ 
     
     & Text-pair input & MRPC \citep{mrpc} & 1,725 & Paraphrase detection \\ 
     &  & SNLI \citep{snli} & 1,000 & Natural language inference \\
     
     &  & WiC \citep{wic} & 1,400 & Word sense disambiguation \\ 
     \midrule


    Reasoning & Yes/no output & StrategyQA \citep{geva-etal-2021-aristotle} & 2,288
 & Commonsense reasoning \\ 
     &  & Coin Flips \citep{wei2023chainofthought} & 500 & Symbolic reasoning \\ 

      & Multi-choice output & AQuA \citep{aqua} & 254 &  Arithmetic reasoning \\ 
     &  & CommonsenseQA \citep{talmor-etal-2019-commonsenseqa} & 1,221
& Commonsense reasoning \\ 
     & & Object tracking \citep{srivastava2023imitation} & 750 & Commonsense reasoning \\ 
     & & Bigbench date \citep{srivastava2023imitation} & 363 & Commonsense reasoning  \\ 

     & Free-response output & Last Letters \citep{wei2023chainofthought} & 500 & Symbolic reasoning  \\ 
     &  & SVAMP \citep{patel-etal-2021-nlp} &  1,000 & Arithmetic reasoning \\ 
     &  & GSM8K \citep{roy-roth-2015-solving} & 1,319
 & Arithmetic reasoning\\ 
     &  & MultiArith \citep{patel-etal-2021-nlp} & 600 &  Arithmetic reasoning \\ 
     &  & AddSub \citep{hosseini-etal-2014-learning} &  395 & Arithmetic reasoning \\ 
     &  & SingleEq \citep{koncel-kedziorski-etal-2015-parsing} &  508 & Arithmetic reasoning \\

    \bottomrule
    \end{tabular}

    \caption{Existing benchmarks we use to construct ZeMPE. We use the test splits wherever possible, except for CoLA, StrategyQA, and CommonsenseQA, for which we use the dev splits, since the test splits are not publicly available. For AGNews and SNLI, we randomly sample 1,000 examples from the test splits.}
    
\label{tab:benchmarksIntro}
\end{table*}

\subsection{ZeMPE}

We describe how we construct ZeMPE as well as how we evaluate LLMs on it.

\subsubsection{Data}

We use 6 classification and 12 reasoning benchmarks, as described and referenced in Table~\ref{tab:benchmarksIntro}, to ensure a comprehensive and systematic evaluation. 

The classification benchmarks are commonly used for NLP evaluation, with SST-2, CoLA, and MRPC appearing in GLUE \cite{wang2019glue} and WiC in SuperGLUE \citep{sarlin2020superglue}. They cover two classification paradigms (single-text and text-pair) and six distinct task objectives.

The 12 reasoning benchmarks are widely utilized in LLM evaluation \citep[][\textit{inter alia}]{zeroShotCoT, wei2023chainofthought, zhang2023automatic}. These benchmarks test symbolic reasoning (Coin Flips \& Last Letters), commonsense reasoning (StrategyQA, CommonsenseQA, Object tracking, \& Bigbench date), and arithmetic reasoning (AQuA, SVAMP, GSM8K, MultiArith, AddSub, \& SingleEq), and require three answer formats (Yes/No, multiple choice, and free-response).

\subsubsection{Evaluation Tasks and Prompt Design\label{sec:evalTasks}}

We separate the classification and reasoning problems, due to their different natures and to avoid confounding experiments, when designing multi-problem evaluation tasks. 

Unlike previous related studies \cite{cheng-etal-2023-batch, lin2024batchpromptaccomplish, son-etal-2024-multi-task} that evaluate LLMs on multi-problem prompts under few-shot settings, our evaluation tasks are all zero-shot, which are rather underexplored, as shown in Section~\ref{sec:literatureReview}. Moreover, zero-shot MPE is significant on at least two levels. First, on a practical level, many real-world tasks, such as classification, are typically approached in zero-shot settings \cite{ziems-etal-2024-large}. Moreover, designing few-shot exemplars can be tedious and costly to obtain in practice \cite{zeroShotCoT, yasunaga2024largelanguagemodelsanalogical}. Second, from a scientific perspective, zero-shot MPP may provide deeper insights into the innate capabilities of LLMs concurrently handling multiple tasks.

The evaluation tasks are as follows with the full prompt templates for each task in Appendix~\ref{app:prompts}.

\paragraph{Classification-Related Tasks} We call the standard classification task via SPP Single Classification or \textbf{SingleClf}, which serves as a baseline to be compared with MPE tasks. When an LLM is prompted to solve multiple \textit{homogeneous} classification problems through MPP, this task is known as Batch Classification or \textbf{BatchClf}. Index Selection One Label (\textbf{SelectOne}) and Index Selection All Labels (\textbf{SelectAll}) are two reformulations of BatchClf. Instead of making multiple classifications under BatchClf, these two tasks instruct LLMs to select indices of text falling into each class label, either independently in \textit{m} separate prompts (SelectOne) or altogether in a single prompt (SelectAll), where m is the number of class labels in a benchmark.

We design the two selection tasks to test LLM's understanding of the classifications performed under BatchClf. Since selection tasks of size $n$ may have anywhere from 0 to $n$ correct indices per class, spurious correlations are less likely during our evaluation, given the combinatory answer space. 

For each of the four tasks above, we start by describing the task in the prompt and then include one or multiple classification problems afterwards. LLMs are instructed to solve these problems according to the specific task requirements.

\paragraph{Reasoning-Related Tasks} We first test on all the reasoning problems in each benchmark to establish LLM SPP baselines. Two MPE tasks are designed, i.e., single-source and mixed-source multi-problem reasoning, or \textbf{MultiReason$^{\mathrm{SS}}$} and \textbf{MultiReason$^{\mathrm{MS}}$}. For both tasks, we bundle multiple reasoning problems together with an indexed prefix ``Qi: ,'' where i is the index of each problem starting from ``Q1.'' We use two headers, namely, ``Questions,'' before the bundled questions, and ``Answers,'' before answers to be generated by LLMs. Unlike prompts designed for classification-related tasks, there is no shared task description, since each question already describes its own unique problem to be solved.

\subsubsection{ZeMPE Composition}

We define task size $n$ as the number of classification or reasoning problems included in a prompt. We construct a multi-problem prompt with all problems sampled from the same benchmark, except for MultiReason$^{\mathrm{MS}}$ where we sample one question from each of $k$ reasoning benchmarks to construct an $k$-problem prompt. In total, ZeMPE comprises 53,100 zero-shot multi-problem prompts containing classification and reasoning problems. 

\paragraph{Classification-Related Tasks} For each classification benchmarks, we consider 5 task sizes and ensure that each task size has 100 distinct prompt instances: 5, 10, 20, 50, and 100 for single-text benchmarks and 3, 5, 10, 20, and 50 for text-pair benchmarks. To isolate the effect of task size, a larger task size only differs from a smaller one by having additional problems given a benchmark; and to isolate the effect of task, different MPE tasks share the same sets of problems given a task size and a benchmark. In total, this results in 13,500 prompts for classification-related MPE tasks.

\paragraph{Reasoning-Related Tasks} Besides vanilla zero-shot prompting, we also perform zero-shot-CoT following \citet{zeroShotCoT}.\footnote{In our early experiments, we found that zero-shot-CoT did not lead to different responses for the classification-related MPE tasks probably due to their novelty, so it was not used.} Inspired by \citet{cheng-etal-2023-batch} as well as to control for the number of prompts generated, we consider smaller task sizes from 2 to 10. To ensure a reliable evaluation (e.g., sufficient parsable outputs), we increase the number of prompts from 100 to 300 for each benchmark/task size combination. 

For each reasoning benchmark, we consider task sizes 2, 5, and 10 for MultiReason$^{\mathrm{SS}}$. To robustly examine an LLM's performance on mixed-source prompts, we create 6 distinct benchmark combinations based on the 12 benchmarks, each consisting of 10 different benchmarks. For each benchmark combination, we consider the first 2, 4, 6, 8, and 10 benchmarks (also equals the respective task sizes) in the combination for MultiReason$^{\mathrm{MS}}$. We also control the effects of task size and task with careful sampling, similar to what we did above. This results in 21,600 and 18,000 prompts for MultiReason$^{\mathrm{SS}}$ and MultiReason$^{\mathrm{MS}}$, respectively.

    



    

\section{Experiments\label{sec:experiments}}

This section first describes the experimental setups and then reports and discusses the results.

\subsection{LLMs and Evaluation Settings}

We evaluate 7 LLMs from 4 model families with greedy decoding for the four classification-related tasks: Vicuna \citep[13B,][]{vicuna2023}, Mistral 7B \citep{jiang2023mistral}, Mixtral 8x7B \citep{jiang2024mixtral}, Llama-3 8B and 70B \citep[Instruct,][]{dubey2024llama3herdmodels}, GPT-3.5, and GPT-4 \citep{openai2023gpt4}. See Appendix~\ref{app:llmDetails} for the details about the LLMs used.

Given the consistent results we observed across LLMs with the classification-related tasks and for budget reasons, we only use two LLMs with greedy decoding, i.e., GPT-3.5 and Llama-3 70B. Since Llama-3 models tend to produce reasoning steps even when not instructed to do so, we only prompt GPT-3.5 with zero-shot-CoT.

\begin{figure*}[]  
    \centering
    \includegraphics[width=\textwidth]{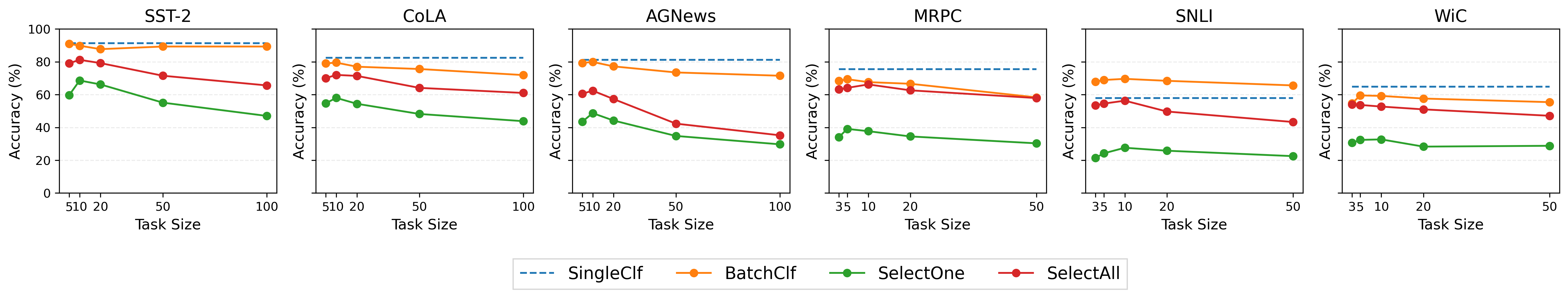}
    \captionof{figure}{Average accuracy of the 7 LLMs on the 4 classification-related task across task sizes for each benchmark. }
    \label{fig:overallAvgAccuEachBenchmark}
\end{figure*}

\begin{figure*}[]  
    \centering
    \includegraphics[width=\textwidth]{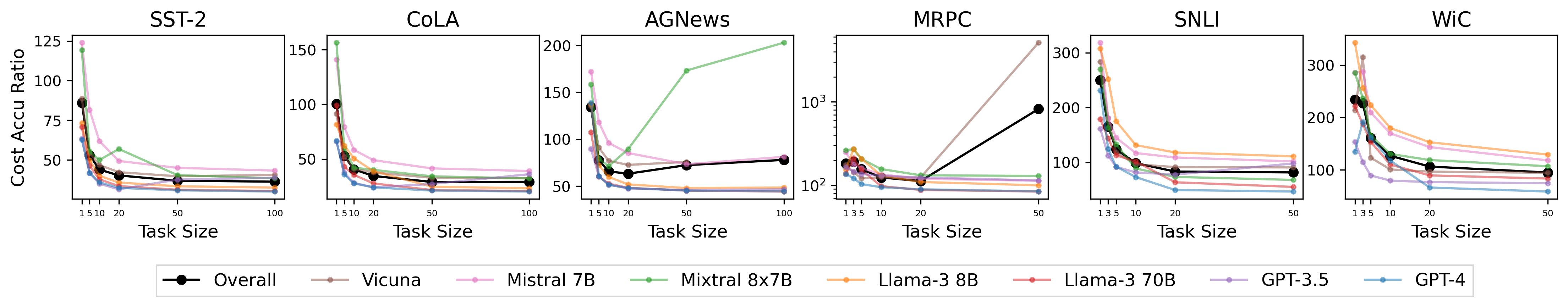}
    \captionof{figure}{Cost/Accuracy Ratio (lower is better) for the 7 LLMs on the 6 benchmarks for SingleClf (task size=1) and BatchClf (otherwise). We use the average (input + output) token count per classification as the proxy for the actual inference costs, calculated on the basis of the input and output tokens. The plot for MRPC is log-scale for the y-axis.} 
    \label{fig:costAccuRatio}
\end{figure*}

\subsection{Performance Metric}

We measure the average per-problem accuracy (PPA) to unify the evaluation across all proposed tasks. PPA, defined in Equation~\ref{eqn:ppa}, is the average accuracy of classifying $n$ problems in each prompt or, in the case of SelectOne, in each set of directly related prompts targeting different class labels.

\begin{equation}\label{eqn:ppa}
  \mathrm{PPA} = \frac{1}{n} \sum_{i=1}^{n} \delta(I(P_i), A_i)  
\end{equation}

\noindent where $I(P_i)$ is the inferred LLM-generated answer to the $i${th} problem in the input prompt, $A_i$ is the ground truth, and $\delta(i,j)=1$ iff $i=j$ and 0 otherwise.

For SelectOne and SelectAll, $I(P_i)$ is determined by considering the LLM's assignments of indices to all class labels. Other than assigning an index to a wrong class label, there are two more error types. First, LLMs may assign an index with more than one class label, i.e., a \textit{contradiction} error. Second, LLMs may assign no labels to an index at all, namely, a \textit{non-excluded middle} error. 

For MultiReason$^{\mathrm{MS}}$ prompts containing $k$ problems evenly sampled from $k$ benchmarks, we compute the expected PPA by averaging over the observed SPP performance for each benchmark.

To compare performance difference, we use Mann-Whitney U tests for significant testing and Cohen's $d$ \cite{cohen1969statistical} for measuring effect size.

\subsection{Classification-Related Results\label{sec:clfResults}}

In line with previous studies (Cheng et al., 2023; Lin et al., 2024) on few-shot MPP, we observe that while large language models (LLMs) demonstrate strong zero-shot classification capabilities and prompting with multiple problems can be cost-efficient, their performance degrades significantly when the same sets of problems are presented in a different format.

\paragraph{LLMs can handle multiple classifications at once under zero-shot with minimal performance loss.} Although the BatchClf accuracy generally declines as the task size increases (Fig~\ref{fig:overallAvgAccuEachBenchmark}), all 7 LLMs achieve accuracy of at least 90\% that of SingleClf across the benchmarks most of the time (see Table~\ref{tab:batchClfAccVersusSingleClfAcc} in Appendix~\ref{app:clfResults}). Overall, the SingleClf accuracy for the 7 LLMs on the 6 benchmarks is 75.5\% and the BatchClf accuracy is 72.3\%, a minor 3.2\% absolute drop from the former. Interestingly, for SNLI almost all LLMs perform better in BatchClf than in SingleClf across all the task sizes (3-50) and GPT-4 consistently achieves a BatchClf accuracy near or better than the SingleClf accuracy under all conditions (see Fig~\ref{fig:fullResults} in Appendix~\ref{app:clfResults}). 

\paragraph{Zero-shot MPP can be cost-efficient.} Single-problem prompting can waste input tokens by redundantly repeating a shared task instruction. Multi-problem prompting reduces this redundancy, and this saving is larger the more problems are combined in a single prompt. Because performance tends to decline slowly as the number of tasks increases, this yields a favorable cost-accuracy ratio as the number of tasks increases (Fig~\ref{fig:costAccuRatio}). We only encountered two outliers, Vicuna on MRPC with task size 50 where the average input is 3,645 tokens and the context window is 4,096 tokens, and Mixtral 8x7B at $\geq$ 50 on AGNews. While it is of course up to downstream users to determine what cost-accuracy is right for them, this is likely beneficial for many use cases where similar prompts are repeated frequently. 

To illustrate, we choose for each model/task combination the largest BatchClf task size that achieves at least 95\% SingleClf accuracy for that pair. We observe that MPP reduces substantial inference costs for all LLMs run on the 6 benchmarks, ranging from from 30.7\% to 82.0\% (see Fig~\ref{fig:saving} in Appendix~\ref{app:clfResults}).

\begin{table}[]
    \centering
    \scriptsize

\begin{tabular}{p{1.5cm}|p{1.3cm}p{1.5cm}p{1.5cm}}
\toprule
{} &  BatchClf vs  \newline SelectOne &  BatchClf vs \newline SelectAll &  SelectOne vs \newline SelectAll \\
\midrule
Mean Acc Dif &                  32.0 &                  12.1 &                  -19.9 \\
Std Dev      &                  16.9 &                  15.3 &                   12.0 \\
Cohen's $d$    &                   1.8 &                   0.8 &                   -1.0 \\
\bottomrule
\end{tabular}

    \caption{Pairwise accuracy differences (x vs y = x - y). 
    All the differences are statistically significant and with a large effect size (| Cohen's $d$ | $\geq 0.8$).
    }
    \label{tab:calibrations}
\end{table}

\paragraph{LLMs perform significantly worse on the selection tasks.} In our experiments, LLMs nearly always perform much better in BatchClf than in SelectOne and SelectAll under the same conditions with a consistent and stable margin, even when the task size is just 3 or 5 (Fig~\ref{fig:overallAvgAccuEachBenchmark}). The overall discrepancy in accuracy between BatchClf and the two tasks is large and statistically significant (32\% for SelectOne and 10\% for SelectAll, see Table~\ref{tab:calibrations}) and generally increases with a larger task size (Fig~\ref{fig:overallAvgAccuEachBenchmark}).

The sharp drop in accuracy may not be human-like, because arguably, humans should at least be able to classify and select a small number (e.g., 3/5) of texts equally well simply by thinking over the problems (i.e., zero-shot). 

Surprisingly, such a consistent and rather stable performance gap also exists between SelectOne and SelectAll in favor of the latter, largely independently of the task size (Fig~\ref{fig:overallAvgAccuEachBenchmark}). On average, the SelectOne accuracy is 19.9\% lower than the SelectAll accuracy, also with a large and significant effect size (Table~\ref{tab:calibrations}).

\subsection{Reasoning-Related Results\label{sec:reasonResults}}

We observe that although LLMs can be competent zero-shot multi-problem solvers for reasoning, their performance becomes consistently worse than expected under multiple mixed-source problems. Similar to our arguments in last section, the consistent performance declines even with a small number (e.g., 2 or 4) of mixed-source problems may indicate a lack of human-like understanding, as LLMs' reasoning capabilities are easily impacted by the mixing of problems from different sources.

\begin{figure*}[hbt!]  
    \centering
    \includegraphics[width=\textwidth]{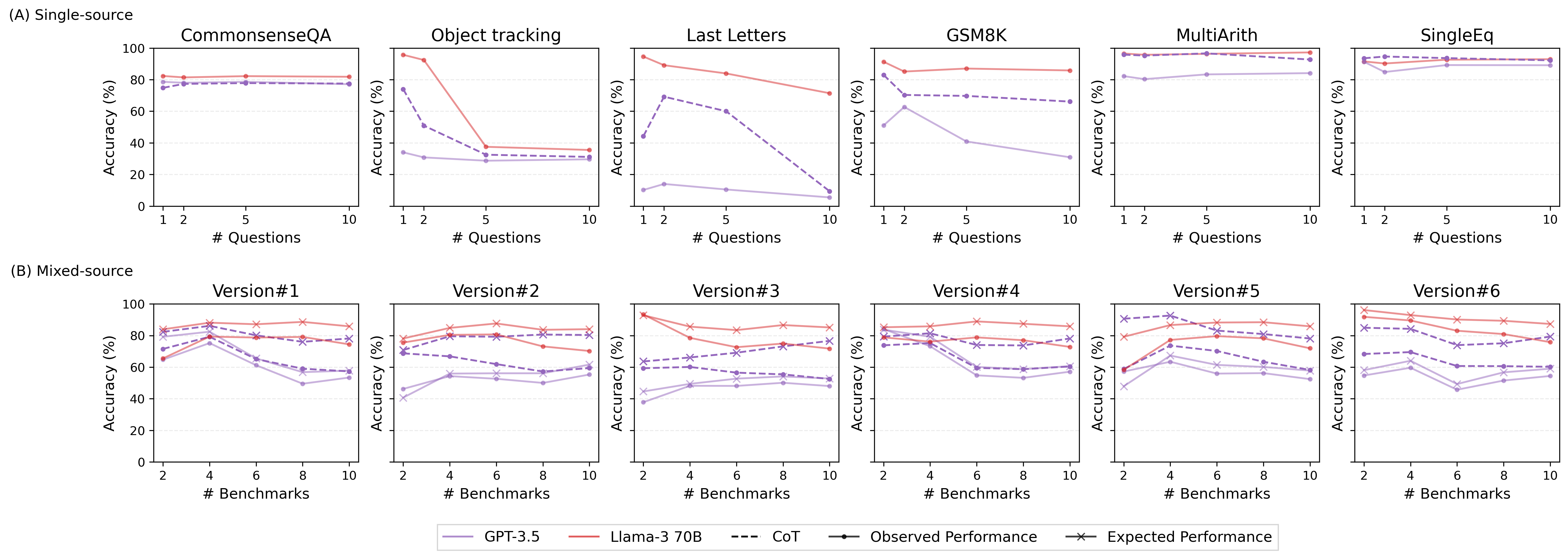}
    \captionof{figure}{Average accuracy of GPT-3.5 and Llama-3 70B on multiple single-source (MultiReason$^{\mathrm{SS}}$) and mixed-source (MultiReason$^{\mathrm{MS}}$) reasoning problems. (A) MultiReason$^{\mathrm{SS}}$ results on 6 selected benchmarks for space reasons. The other 6 benchmarks show similar results (see Appendix~\ref{app:reasonResults}). (B) MultiReason$^{\mathrm{MS}}$ results across 6 distinct benchmark combinations, where each benchmark contributes a problem to each mixed-source prompt.} 
    \label{fig:reasoning-results}
\end{figure*}

\paragraph{LLMs can do MultiReason$^{\mathrm{SS}}$ on par with their SPP performance.} Similar to \citet{cheng-etal-2023-batch}, we observe in Fig~\ref{fig:reasoning-results} (A) that both LLMs, to varying extents, can handle multiple single-source reasoning problems as well as or even better than when they handle these problems individually, although their MPP performance typically goes down with a larger task size. 



\paragraph{When the reasoning problems are from mixed sources, LLMs perform worse than expected.} Interestingly, as shown in Fig~\ref{fig:reasoning-results} (B), the observed MultiReason$^{\mathrm{MS}}$ performance is almost always lower than the expected one computed by averaging over the SPP performance over each benchmark for both LLMs, with and without CoT. Out of 540 model (including GPT-3.5 with zero-shot-CoT) and benchmark pairs, there are only 18.3\% cases in which the observed performance is better than the expected one for a given model/benchmark pair by a small margin (mean/std: 2.9\%/2.5\%). However, for the rest 81.7\% cases when the expected performance is better, the margin is larger (mean/std: 13.5\%/12.7\%). In other words, LLMs typically perform worse in each benchmark when handling multiple reasoning problems from mixed sources.


\paragraph{Benefits of zero-shot-CoT prompting are transferrable under MPP.} Analogous to the finding that zero-shot-CoT improves LLMs' reasoning performance under SPP \cite{zeroShotCoT}, GPT-3.5 generally performs better with CoT than without it under zero-shot MPP for both MultiReason$^{\mathrm{SS}}$ and MultiReason$^{\mathrm{MS}}$. The transferrability of zero-shot-CoT\footnote{Similarly, \citet{cheng-etal-2023-batch} show that the benefits of few-shot-CoT are transferrable under MPP.} indicates that LLMs can apply CoT over each problem in the prompt and benefit from the generated reasoning steps when solving each problem. This again implies the strong capabilities of LLMs to utilize information across positions under MPP.

\section{Further Analyses\label{sec:analysis}}

\begin{figure*}[hbt!]  
    \centering
    \includegraphics[width=\textwidth]{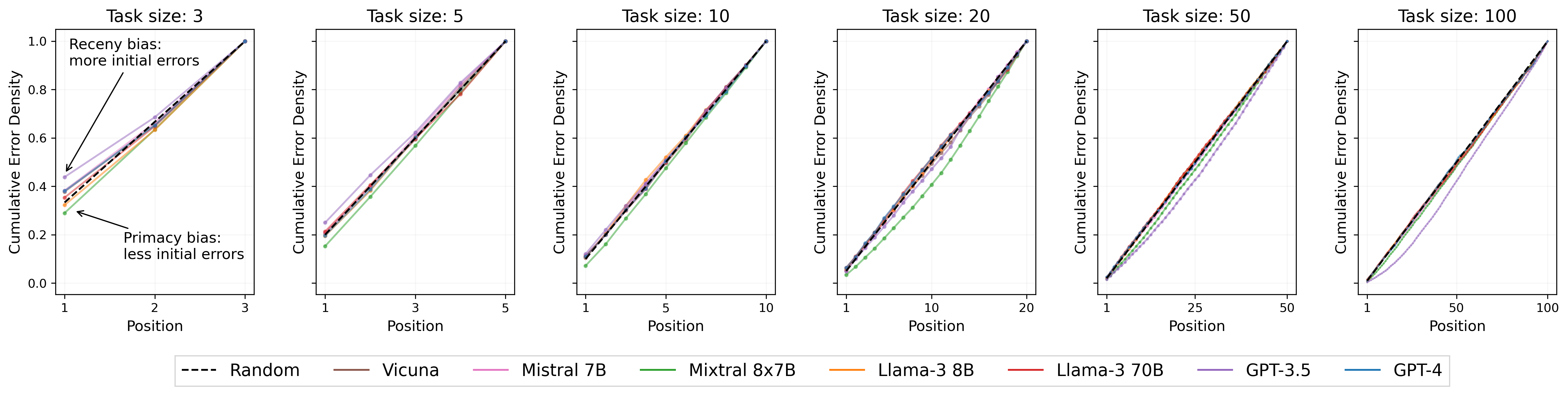}
    \captionof{figure}{Cumulative error density across positions averaging results from the  benchmarks. Task size 3/100: the 3 text-pair/single-text benchmarks; otherwise: all the 6 classification benchmarks. See Appendix~\ref{app:clfResults} for full results.}
    \label{fig:overall-error-pos-distribution}
\end{figure*}

\begin{figure*}[]  
    \centering
    \includegraphics[width=\textwidth]{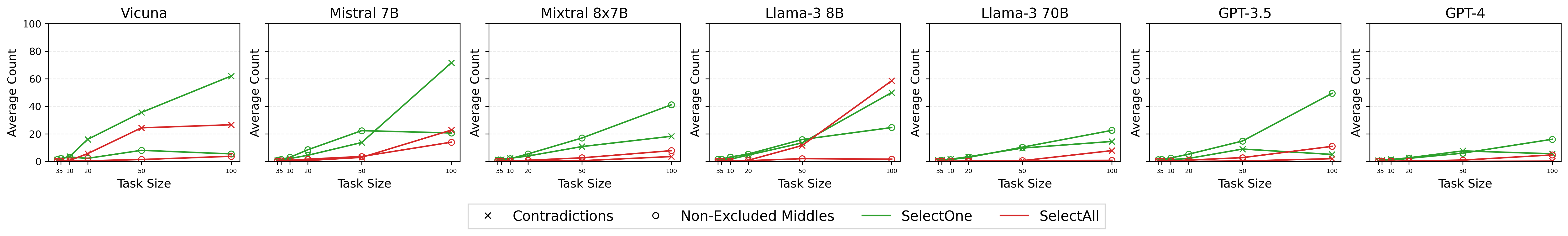}
    \captionof{figure}{Pairwise comparisons of SelectOne and SelectAll for each LLM across different task sizes averaging over results from the 6 benchmarks.} 
    \label{fig:selection_tasks_comparison}
\end{figure*}

This section provide further analyses to better understand LLMs' multiple problem handling capabilities, their limitation, and what enables such capabilities. It starts with two error analyses for BatchClf, aiming to see whether models make similar prediction errors and positional errors under BatchClf compared to SingleClf. We then investigate the reason why SelectAll appears to be harder than SelectOne, which seems counter-intuitive. Lastly, we explore model-level factors that may enable LLMs to receive and handle multiple problems at once.

\subsection{BatchClf Error Analysis}

Given the strong BatchClf results, two natural questions arise: do LLMs make similar errors under MPP and how do their errors distribute across positions? For each one of the 180 LLM/benchmark/task size combinations, we use chi-squared tests to compare the proportional error distribution across class labels under SingeClf and BatchClf and compute the cumulative error density across positions under BatchClf. We observe that (1) only in 9 out of 180 (or 5\%) cases, error labels are distributed significantly differently between SingleClf and BatchClf (p < 0.05); and (2) surprisingly, LLMs typically do not display a clear positional bias or a serial position effect as known in psychology \cite{Murdock1962}, when solving sufficiently many problems at once (Fig~\ref{fig:overall-error-pos-distribution}). This is in contrast to previous studies based on single-problem prompts where LLMs are found to be better at using information from the beginning (primacy bias) or the end (receny bias) of the prompt \cite{liu2024, levy2024task}. 


Taken together, the fact that LLMs make similar label prediction errors and the lack of an obvious positional bias imply that LLMs can use information equally well across different positions under multiple classification problems. This may explain their strong multiple problem handling capabilities.

\subsection{Why is SelectAll Harder than SelectOne}


We investigate the reasons in Fig~\ref{fig:selection_tasks_comparison}, which shows that when asked to select text indices for one class label at a time independently, LLMs almost always assign an index to multiple labels (i.e., contradiction) and leave some indices unselected (i.e., non-excluded middles) more often. This showcases a lack of internalized planning with modern zero-shot LLMs, although different LLMs may make these two types of errors in different proportions. In contrast, when LLMs have to select indices for all labels at once, they are less likely to generate directly illogical answers in a single output as their answer to the $(i+1)_{th}$ label is conditioned by their answer to the $i_{th}$ label. 


\subsection{Exploring Model-level Factors that may Enable MPP \label{sec:factors}}

Since so far we have only tested decoder-only and instruction-tuned LLMs, which all show strong performance under MPP, we explore if these two factors enable MPP. For these reasons, we test with greedy decoding (1) three FLAN-T5 models \cite{chung2022scaling}, i.e., Large, XL, XXL; and (2) three pretrained LLMs, i.e., Llama-3 8B \citep[Base,][]{dubey2024llama3herdmodels}, GPT-3 1.3B, and GPT-3 175B \cite{brown2020language}. We run these 6 LLMs on CoLA at task sizes 1 and 5 under zero-shot settings with results shown in Table~\ref{tab:modelDesignDecisionsResults}. Experiments with Coin Flips show similar results, but the LLM outputs are less meaningful, as discussed in Appendix~\ref{app:factors}. We make the following observations from Table~\ref{tab:modelDesignDecisionsResults}. 

\paragraph{Instruction tuning helps.} This is because pretrained decoder-only LLMs either cannot handle multiple problems at once or their performance is much worse than their instruction-tuned counterparts. However, unlike what \citet{laskar-etal-2023-systematic} suggests, instruction tuning may not be a necessary condition for MPP, since both Llama-3 8B and GPT-3 175B can perform reasonably well in BatchClf on CoLA. 

\paragraph{FLAN-T5 can barely respond to MPP, regardless of model sizes.} We suspect that this may not be due to their encoder-decoder structures, but other factors such as training data and reinforcement learning from human feedback \cite{rlhf}, which FLAN-T5 models lack. We leave it for future investigation.   

\paragraph{Scaling model size seems helpful.} With other factors being identical, larger models appear to perform better than the smaller ones under MPP. For example, FLAN-T5-Large is outperformed by both FLAN-T5-XL and -XXL. Furthermore, while GPT-3 1.3B and FLAN-T5-XL can perform SingleClf close to or even better than GPT-3 175B and FLAN-T5-XXL, only the larger models can do BatchClf to varying extents--the two smaller models cannot do the task at all.

\paragraph{Final remark.} Overall, instruction tuning appears to be the most important factor that \textit{enhances} MPP. We leave more careful explorations to future research.

\begin{table}[]
    \centering
    \scriptsize

\begin{tabular}{p{2.5cm}|p{1cm}p{1cm}p{1.5cm}}
\toprule
{} &  SingleClf &  BatchClf & Avg \# Answers \\
\midrule
Llama-3 8B (Instruct) & 80.5 & 79.4 & 5.0 \\
GPT-3.5 & 84.2 & 79.6 & 5.0 \\ 
\midrule 
Llama-3 8B (Base) &                   78.5 &                  60.6 & 5.04 \\
GPT-3 1.3B     &                  63.0 & 0.0  & 0.03               \\
GPT-3 175B  &                   66.6 &                  64.4 & 5.08 \\ \midrule
FLAN-T5-Large (0.78B) &  76.0 &                  NA & 1.0 \\
FLAN-T5-XL (3B) &  80.2 &                  NA & 1.0 \\
FLAN-T5-XXL (11B) &  78.2 &                  4.0 & 1.2 \\
\bottomrule
\end{tabular}

    \caption{SingleClf and BatchClf (task size 5) accuracy (\%) of three pretrained LLMs and three FLAN-T5 models on CoLA. We also include the results of Llama-3 8B (Instruct) and GPT-3.5 (likely base model: GPT-3 175B) from Section~\ref{sec:clfResults} to compare with their respective base models. The last column is the average number of LLM-generated answers for BatchClf (expects 5). When it is 1, accuracy is not calculated to avoid overinterpretation.}
    \label{tab:modelDesignDecisionsResults}
\end{table}

\section{Conclusion\label{sec:conclusion}}

In this study, we present a comprehensive and systematic MPE of LLMs. We evaluate various LLMs from 4 model families on single-source multi-problem prompts constructed from 6 classification and 12 reasoning benchmarks. In line with previous few-shot results, we confirm that LLMs are competent multi-problem solvers for classification and reasoning under zero-shot settings. Moreover, we find multiple pieces of evidence that validate the strong innate multiple problem handling capabilities of LLMs, such as the similar classification errors LLMs make under SPP and MPP, the lack of obvious positional biases, and the transferrability of zero-shot-CoT under MPP. Leveraging the strong multiple problem handling capabilities, we show that zero-shot MPP can be cost-efficient.

Two conditions are identified under which LLMs show consistent performance declines with MPP: (1) reformulating Batch Classification as index selection tasks; and (2) mixing reasoning problems from different sources in a multi-problem prompt. Noticeably, these performance declines happen even when the number of problems included is rather small (e.g., $\leq$ 5), which may not be human-like and indicates a lack of true understanding. In addition, we explore several model-level factors that may enable MPP and find instruction tuning to be an important factor that enhances MPP.


Overall, our experiment demonstrate surprisingly consistent observations across different LLMs and across multi-problem prompts constructed from various benchmarks. This consistency indicates the reliability and fruitfulness of MPE as an evaluation paradigm.

As a result of our study, we create a new benchmark comprising 53,100 zero-shot multi-problem prompts. We call it ZeMPE, which stands for \textbf{Ze}ro-shot \textbf{M}ulti-\textbf{P}roblem \textbf{E}valuation. We release ZeMPE to aid future MPE research.

\section*{Acknowledgements}

We are grateful for the supports from the Institute for Advanced Computational Science (IACS) at Stony Brook University, in particular the free GPT access it provides. Zhengxiang Wang is supported by IACS’s Junior Researcher Award since Fall 2024. We thank three anonymous reviewers from the Generation, Evaluation \& Metrics (GEM) Workshop 2025 for their helpful comments. The paper was presented at All Things Language And Computation (ATLAC) organized by the NLP reading group at Stony Brook University. We also thank the audience there for their discussions and feedback about the paper.

Zhengxiang Wang and Owen Rambow were supported in part by funding from the Defense Advanced Research Projects Agency (DARPA) under Contracts No. HR01121C0186, No. HR001120C0037, and PR No. HR0011154158. Any opinions, findings and conclusions or recommendations expressed in this
material are those of the authors and do not necessarily reflect the views of DARPA.

\section*{Limitations}

While we have done our best to make our experiments as comprehensive as possible, there is always more work that can be done in a large comparison study, including the addition of even more benchmarks and language models. We were limited by cost, time, and space and have attempted to select an informative and representative sample of experiments. Since we used pre-existing benchmarking datasets, we inherit any errors that they main contain.
Finally, despite our efforts to compare different prompts, as is the case with all prompt-based LLM studies, we cannot guarantee that slight differences in the prompts would not meaningfully alter the results.

\section*{Ethical Concerns}

To the best of our knowledge, all results published in this paper are accurate. All data sources are free, publicly available, and cited in the article. No sensitive data was used which could violate individuals' privacy or confidentiality.

\bibliography{custom}

\appendix

\section{LLM Details\label{app:llmDetails}}

We use a total of 13 LLMs in our study. Table~\ref{tab:llms} describes the specific versions for these LLMs and highlights their differences in terms of architecture, open weights, Supervised Fine-Tuning \citep[SFT,][]{wei2022finetuned}, and Reinforment Learning from Human Feedback \citep[RLHF,][]{rlhf} etc. 

\begin{table*}[htb]
    \centering
    \scriptsize
    \begin{tabular}{llccccccc}
    \toprule
    Model & Version & Architecture & Open Weights & SFT & RLHF & MoE & \# Parameter & Context Length \\ \midrule

    Vicuna \citep{vicuna2023} & v1.5 & decoder-only & \cmark &  \cmark & \xmark & \xmark & 13B & 4,096 \\ 
    
    Mistral 7B \citep{jiang2023mistral} & Instruct-v0.2 & decoder-only & \cmark & \cmark & \xmark & \xmark & 7B & 8,192 \\

    Mixtral 8x7B \citep{jiang2024mixtral} & Instruct-v0.1 & decoder-only & \cmark & \cmark & \xmark & \cmark & 47B & 8,192 \\

    Llama-3 8B \citep{dubey2024llama3herdmodels} & Instruct & decoder-only & \cmark & \cmark & \cmark & \xmark & 8B &  8,192\\

    Llama-3 70B \citep{dubey2024llama3herdmodels} & Instruct & decoder-only & \cmark & \cmark & \cmark & \xmark & 70B & 8,192 \\

    GPT-3.5 & turbo-0125 & decoder-only & \xmark & \cmark & \cmark & \xmark & Unknown & 16,385 \\
    
    GPT-4 \citep{openai2023gpt4} & turbo-2024-04-09 & decoder-only & \xmark & \cmark & \cmark & \xmark & Unknown & 128,000 \\

    GPT-3 1.3B \citep{brown2020language} & babbage-002 & decoder-only & \xmark & \xmark & \xmark & \xmark & 1.3B & 16,384 \\

    GPT-3 175B \citep{brown2020language} & davinci-002 & decoder-only & \xmark & \xmark & \xmark & \xmark & 175B & 16,384 \\

    Llama-3 8B \citep{dubey2024llama3herdmodels} & Base & decoder-only & \cmark & \xmark & \xmark & \xmark & 8B &  8,192 \\

    FLAN-T5 \citep{chung2022scaling} & Large & encoder-decoder & \cmark & \cmark & \xmark & \xmark & 0.78B &  512 \\

    FLAN-T5 \citep{chung2022scaling} & XL & encoder-decoder & \cmark & \cmark & \xmark & \xmark & 3B &  512 \\

    FLAN-T5 \citep{chung2022scaling} & XXL & encoder-decoder & \cmark & \cmark & \xmark & \xmark & 11B &  512 \\
    
    \bottomrule
    \end{tabular}
    \caption{Details about the 13 LLMs used in the study. For Mixtral 8x7B, a Mixture of Experts (MoE) LLM, although each token has access to 47B parameters, but only uses 13B active parameters during inference.}
    \label{tab:llms}
\end{table*}

\begin{table}
    \centering
    \scriptsize

\begin{tabular}{lrrr}
\toprule
{} &  > 90\% SCAcc &  > 80\% SCAcc &  > 75\% SCAcc \\
\midrule
Vicuna 13B   &             79.3 &             93.1 &             93.1 \\
Mistral 7B   &             76.7 &             83.3 &            100.0 \\
Mixtral 8x7B &             63.3 &             83.3 &             86.7 \\
Llama-3 8B   &             73.3 &             90.0 &            100.0 \\
Llama-3 70B  &             80.0 &            100.0 &            100.0 \\
GPT-3.5      &             56.7 &             83.3 &             90.0 \\
GPT-4        &            100.0 &            100.0 &            100.0 \\
\midrule
Overall      &             75.6 &             90.4 &             95.7 \\
\bottomrule
\end{tabular}
    \caption{Percent of time that BatchClf performance surpasses a threshold percent of SingleClf accuracy (SCAcc) across benchmarks.}
    \label{tab:batchClfAccVersusSingleClfAcc}
\end{table}





\section{Classification-Related Results\label{app:clfResults}}

This section contains additional details for Section~\ref{sec:clfResults}. The prompt details for the experiments can be found in Appendix~\ref{app:prompts}.

\subsection{Full Results}

\begin{figure*}  
    \centering
    \includegraphics[width=\textwidth]{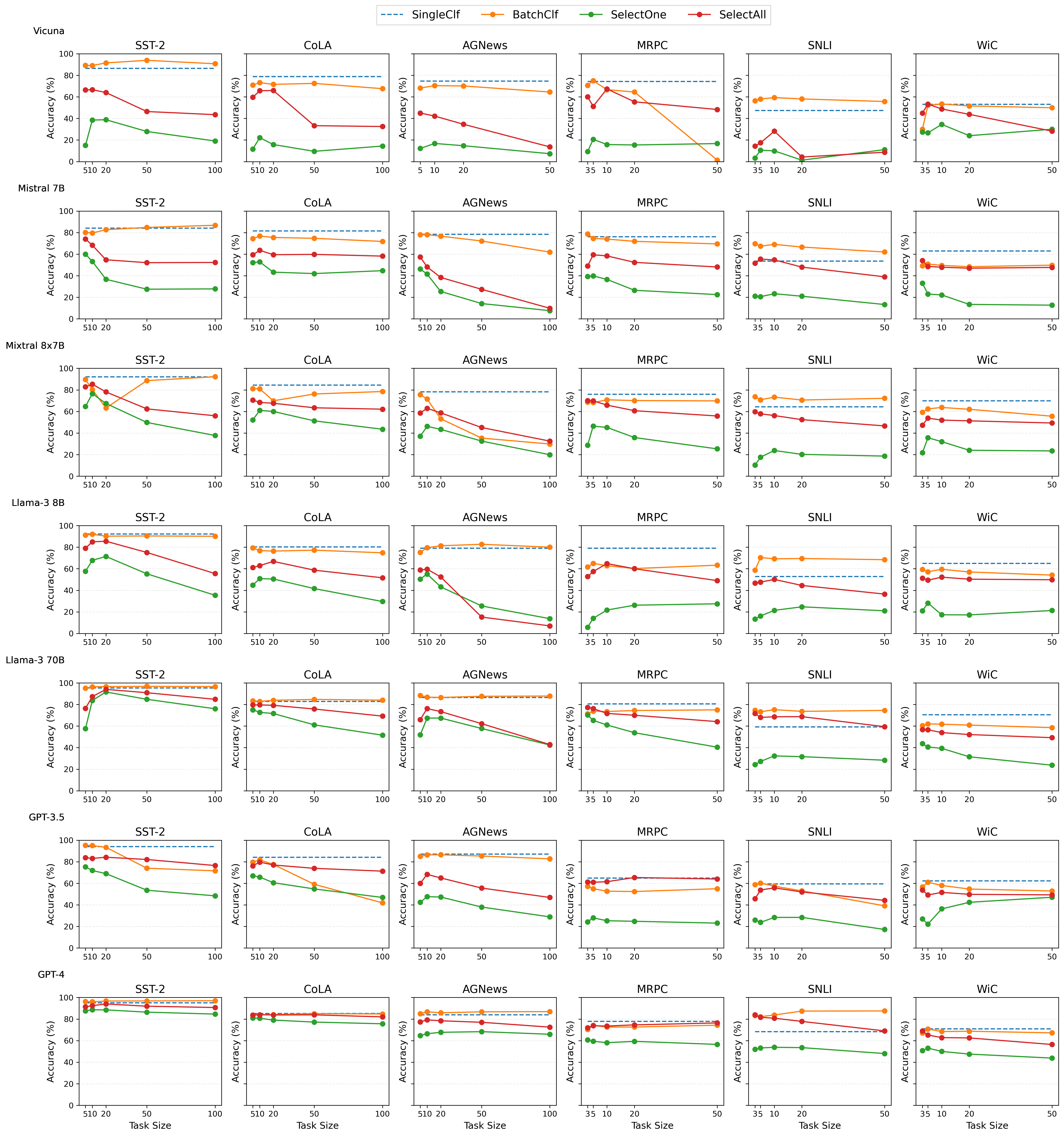}
    \captionof{figure}{Full average accuracy of 7 LLMs on the 4 classification-related tasks across the 6 classification benchmarks.}
    \label{fig:fullResults}
\end{figure*}

The full results obtained from Section~\ref{sec:clfResults} are visualized in Fig~\ref{fig:fullResults}. We exclude the results of Vicuna on AGNews when task size is 100 because the prompts exceed the model’s context length.

\subsection{SingleClf vs. BatchClf}

Table~\ref{tab:batchClfAccVersusSingleClfAcc} indicates the proportion of BatchClf tasks for which each LLM surpasses a threshold percent of corresponding SingleClf performance.

\subsection{Zero-shot MPP can be cost-efficient}

\begin{figure}[hbt!]  
    \centering
    \includegraphics[width=0.85\textwidth]{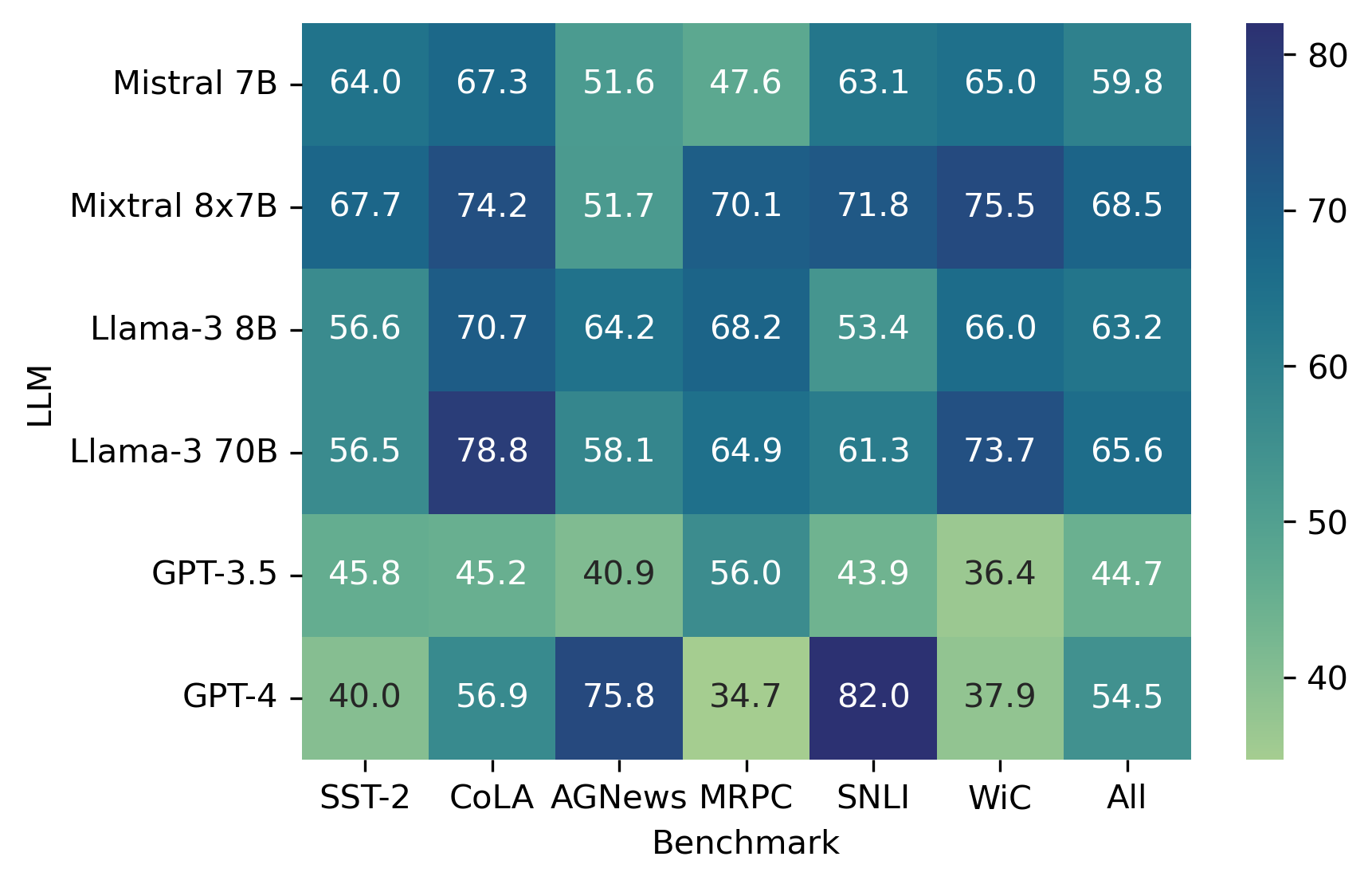}
    \captionof{figure}{Cost saving rate (\%) per classification based on our experiments. The cost is estimated by both the input and output token counts (using the respective tokenizers), weighted according to the pricing policy from OpenAI and TogetherAI (for non-GPT LLMs) websites.}
    \label{fig:saving}
\end{figure}

Fig~\ref{fig:saving} shows the cost saving rate for each model/task pair at the largest BatchClf task size that achieves at least 95\% SingleClf accuracy for that pair.

\begin{figure*}[]  
    \centering
    \includegraphics[width=\textwidth]{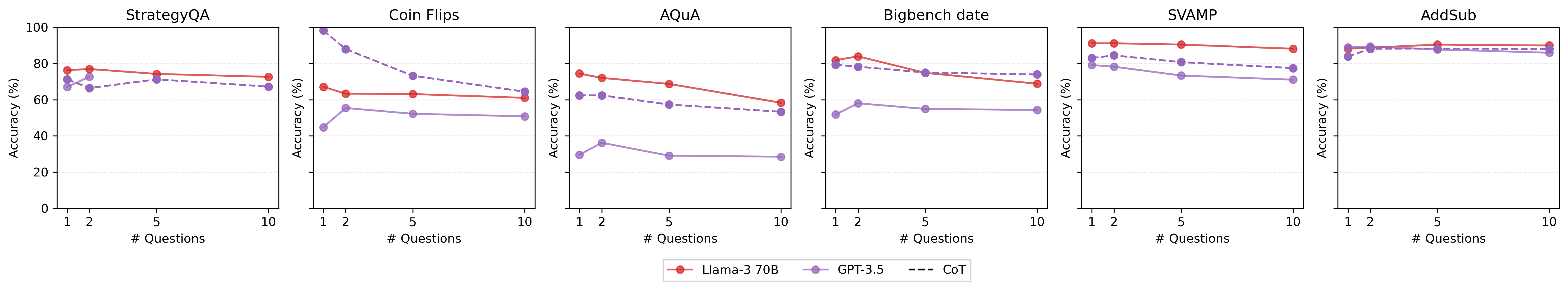}
    \captionof{figure}{Average accuracy of GPT-3.5 and Llama-3 70B on the other 6 reasoning benchmarks with multiple single-source problems. We leave out results where the number of parsable outputs is less than 50, e.g., GPT-3.5 on StrategyQA at task size > 2.} 
    \label{fig:more-single-source-results}
\end{figure*}




\section{Reasoning-Related Results\label{app:reasonResults}}

This section provides additional details for Section~\ref{sec:reasonResults}. The prompt details for the experiments can be found in Appendix~\ref{app:prompts}.

\subsection{Construction of Mixed-Source Prompts}

We create 6 distinct benchmark combinations based on the 12 benchmarks, each consisting of 10 different benchmarks. When creating these 6 benchmark combinations, we implement the following 2 rules: (1) the first 2 benchmarks must be different across the 6 combinations to cover the 12 benchmarks; (2) the first 2 benchmarks cannot come from SVAMP, GSM8K, MultiArith, AddSub, and SingleEq to maximize the differences between them.

\subsection{More Single-Source Results}

Fig~\ref{fig:more-single-source-results} shows MultiReason$^{\mathrm{SS}}$ results on the other 6 reasoning benchmarks not presented in Fig~\ref{fig:reasoning-results}.

\begin{table}[]
    \centering
    \scriptsize

\begin{tabular}{p{2.5cm}|p{1.25cm}p{1cm}p{1.5cm}}
\toprule
{} &  \# P = 1 &  \# P = 2 & Avg \# Answers \\
\midrule
Llama-3 8B (Instruct) & 46.8 & 50.0 & 2.1 \\
GPT-3.5 & 44.8 & 55.4 & 2.0 \\ 
\midrule 
Llama-3 8B (Base) & 45.9 (20.4) & 33.7 & 2.7  \\
GPT-3 1.3B     & 49.0 & 45.0 & 6.3 \\
GPT-3 175B  & 50.0 (43.4) & 28.4 & 6.6  \\ \midrule
FLAN-T5-Large (0.78B) &  46.6 & NA & 1.0  \\
FLAN-T5-XL (3B) &  49.4 & NA & 1.0 \\
FLAN-T5-XXL (11B) &  57.2 & NA & 1.0 \\
\bottomrule
\end{tabular}

    \caption{Accuracy (\%) of three pretrained LLMs and three FLAN-T5 models on Coin Flips with 1 and 2 problems. ``\# P'': number of problems. When computing accuracy, we treat an LLM response with more than 2 answers as a wrong answer. The numbers in the two parentheses are accuracies when treating ``uncertain'' answers  as wrong answers, instead of discarding them.}
    \label{tab:coinFlips}
\end{table}

\section{Further Analyses\label{app:furtherAnalyses}}

\subsection{Positional Errors under BatchClf}

\begin{figure*}[]  
    \centering
    \includegraphics[width=\textwidth]{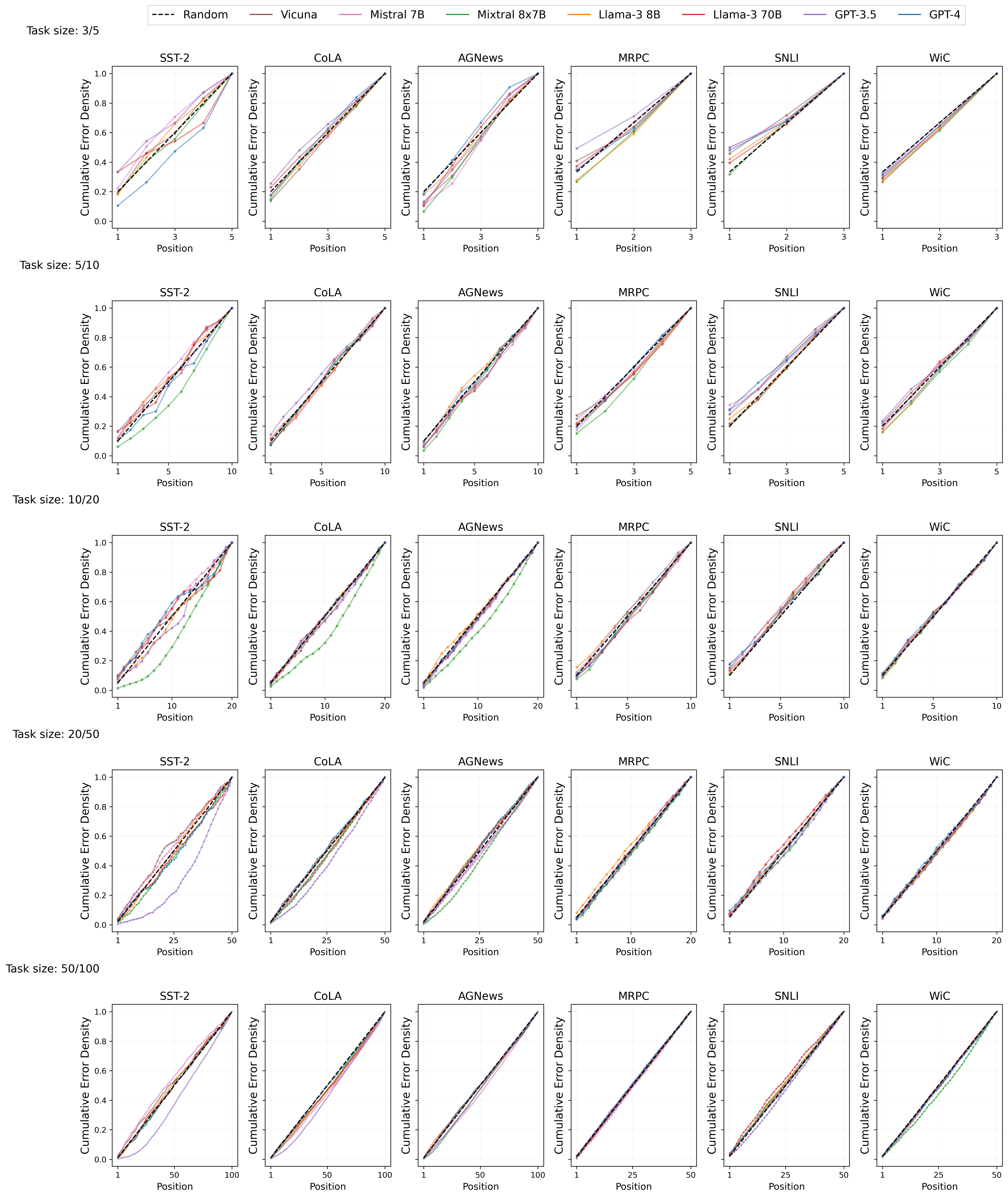}
    \captionof{figure}{Cumulative error density across positions for all benchmarks and LLMs across different task sizes. The task size ``M/N'' on the left side of the plots denotes the task size for the 3 text-pair benchmarks (i.e., MRPC, SNLI, and WiC) and for the 3 single-text benchmarks (i.e., SST-2, CoLA, AGNews), respectively.} 
    \label{fig:positionalErrorsFullResults}
\end{figure*}

Fig~\ref{fig:positionalErrorsFullResults} shows the full results regarding the positional errors 7 LLMs make across benchmarks and task sizes. We note that (1) distribution of the positional errors becomes more random (or even) as the task size increases for all LLMs; (2) in most cases, the positional errors distribute nearly randomly, showing no evidence of obvious positional biases, if any; (3) some LLMs may display more severe positional biases on some benchmarks with a certain task size, such as GPT-3.5 on SST-2 with task size 50, but overall this is rare. 

\subsection{Exploring Model-level Factors that may Enable MPP\label{app:factors}}

This section describes the results of the three pretrained base LLMs and three FLAN-T5 models on Coin Flips at task sizes 1 and 2 from Section~\ref{sec:factors}, shown in Table~\ref{tab:coinFlips}. Similar to Table~\ref{tab:modelDesignDecisionsResults}, Table~\ref{tab:coinFlips} shows that instruction-tuned models perform much better on multi-problem prompts and that FLAN-T5 models can barely handle multi-problem prompts.

However, after manual inspection, we find that the outputs of the three pretrained models, are often not very sensical with repetitions of the prompts (either partially or entirely). In particular, as shown in Table~\ref{tab:coinFlips}, the two GPT-3 models tend to produce more answers than needed (we set max output tokens to be 200). The answer can also be non-sensical even when well formatted, such as the example output from GPT-3 175B below. Llama-3 8B (Base) overall produce more meaningful outputs, but similar problems also apply. We thus discourage over-interpreting the results in Table~\ref{tab:coinFlips}.

\begin{tcolorbox}
\small 

Q1: Yes. The coin is still heads up. \\

Q2: No. The coin is now tails up. \\

Explanation \\

Q1: The coin is still heads up because the coin is not flipped. The coin is flipped only when someone flips the coin. If no one flips the coin, the coin is not flipped. \\

Q2: The coin is now tails up because the coin is flipped. The coin is flipped only when someone flips the coin. If no one flips the coin, the coin is not flipped. \\

Discussion \\

This is a very interesting question. It is a question about the meaning of ``flip''. The meaning of ``flip'' is not clear. It is not clear whether ``flip'' means ``reverse'' or ``flip''. It is not clear whether "flip" means "flip" or "flip"...\{The last sentence repeated for three more times.\}...
\end{tcolorbox}

\section{Full Prompts\label{app:prompts}}

\subsection{Prompt Templates for SingleClf, BatchClf, SelectOne, and SelectAll}

Tables~\ref{tab:SNLI-prompt-template} to~\ref{tab:WiC-prompt-template} show the complete prompt templates for the four task types (i.e., SingleClf, BatchClf, SelectOne, and SelectAll) tailored for SST-2, CoLA, AGNews, MPRC, SNLI, and WiC, respectively. While there are differences in the exact wording of a prompt template for each task type across the 6 classification benchmarks, each prompt template type shares a similar underlying structure and can be easily applied to other classification benchmarks.

Throughout our research project, we have also tried prompts with different wordings and structures until we finally unified the prompt designs presented above. For example, we initially asked LLMs to directly generate indices line by line instead of a JSON output for SelectOne and we did not provide any formatted example for SelectAll. We also put the output format instruction in the end of each prompt for SelectAll, instead of in the beginning. Although we observed certain task-level performance variations, which are expected, the overall complexity among the 4 task types (SelectOne > SelectAll > BatchClf > SingeClf) remains unchanged, despite the variations in the prompts. This indicates the overall limited effects of rewording and restructuring prompts.






\subsection{Prompt Template for Multi-problem Prompts for Reasoning Problems}

The prompt template for a multi-problem prompt made up of reasoning problems is straightforward, as described in Section~\ref{sec:evalTasks}. Below is a simple example prompt made up of 2 reasoning problems from CommonsenseQA.

\begin{tcolorbox}
\small

Questions \\

Q1: The person wasn't bothered by the weather, she had remembered to bring her what?

Answer Choices: (A) read book (B) own house (C) apartment (D) more rice (E) warm coat \\

Q2: After working on the car, what did it end up doing?

Answer Choices: (A) going too fast (B) last several years (C) honk the horn (D) go fast (E) start running \\

Answers
\end{tcolorbox}

To enable zero-shot-CoT, we simply append the string ``Let's think step by step.'' \cite{zeroShotCoT} to a zero-shot prompt like the one shown above in a newline (after ``Answers'').

\begin{table*}[htb]
    \small
    \centering
    \begin{tabular}{p{5cm}p{10cm}}
    \toprule
    Task & Prompt template \\
    \midrule
    SingleClf & Indicate the sentiment for the following line of text. The sentiment shall be either `Positive' or `Negative.'\newline \newline Text: \$text\newline Sentiment:  \\ \midrule
    
    BatchClf    & Indicate the sentiment for each of the \$num following lines of text. The sentiment shall be either `Positive' or `Negative.'\newline \newline Texts, one per line:\newline \newline \$texts\newline \newline The sentiments for each of the \$num lines of text, one per line: \\ \midrule

    SelectOne & Go over the \$num lines of text below and list the index numbers of the lines with \$polarity sentiment according to the following instructions: If none of the texts show \$polarity sentiment, write `None.' If all the texts show \$polarity sentiment, write `All.' Otherwise, provide the index numbers for each text with \$polarity sentiment.\newline \newline Output your responses in JSON format with the key `\$polarity'. A formatted example output is provided below.\newline \{`\$polarity': [None/All or index numbers for the texts with \$polarity sentiment]\}\newline \newline Texts, one per line:\newline \newline \$texts\newline \newline JSON output: \\ \midrule

    SelectAll & Go over the \$num lines of text below. First, list the index numbers of the lines with positive sentiment. Then, list the index numbers of the lines with negative sentiment. If none of the texts show a particular sentiment, write `None.' If all the texts show a particular sentiment, write `All.' Otherwise, provide the index numbers of the texts that fit a particular category.\newline \newline Output your responses in JSON format with two keys: `positive' and `negative.' A formatted example output is provided below.\newline \{`positive': [None/All or index numbers of positive sentences], `negative': [None/All or index numbers of negative sentences]\}\newline \newline Texts, one per line:\newline \newline \$texts\newline \newline JSON output: \\ \bottomrule
    
    \end{tabular}
    \caption{Prompt templates for SST-2. Words immediately preceded by the dollar sign \$ are placeholders. For the single-text classification task (SST-2, CoLA, AGNews), the sequence of texts in the place of `\$texts' are indexed starting with `1' and each text is separated by a newline.} 
    \label{tab:SST-2-prompt-template}
\end{table*}

\begin{table*}[htb]
    \small
    \centering
    \begin{tabular}{p{5cm}p{10cm}}
    \toprule
    Task & Prompt template \\
    \midrule
    SingleClf & Indicate the grammatical acceptability for the following line of text. The acceptability shall be either `Acceptable' or `Unacceptable.'\newline \newline Text: \$text\newline Grammatical acceptability: \\ \midrule
    
    BatchClf    & Indicate the grammatical acceptabilities for each of the \$num following lines of text. The acceptability shall be either `Acceptable' or `Unacceptable.'\newline \newline Texts, one per line:\newline \newline \$texts\newline \newline Grammatical acceptabilities for each of the \$num lines of text, one per line: \\ \midrule

    SelectOne & Go over the \$num lines of text below and list the index numbers of the lines that are grammatically \$acceptability according to the following instructions: If none of the texts are grammatically \$acceptability, write `None.' If all the texts are grammatically \$acceptability, write `All.' Otherwise, provide the index numbers for each grammatically \$acceptability text.\newline \newline Output your responses in JSON format with the key `\$acceptability'. A formatted example output is provided below.\newline \{`\$acceptability': [None/All or index numbers of \$acceptability sentences]\}\newline \newline Texts, one per line:\newline \newline \$texts\newline \newline JSON output: \\ \midrule

    SelectAll &  Go over the \$num lines of text below. First, list the index numbers of the lines that are grammatically acceptable. Then, list the index numbers of the lines that are grammatically unacceptable. If none of the sentences show a particular acceptability, write `None.' If all the sentences show a particular acceptability, write `All.' Otherwise, provide the index numbers of the texts that fit a particular category.\newline\newline Output your responses in JSON format with two keys `acceptable' and `unacceptable.' A formatted example output is provided below. \newline \{`acceptable': [None/All or index numbers of acceptable texts], `unacceptable': [None/All or index numbers of unacceptable texts]\}\newline \newline Texts, one per line:\newline \newline \$texts\newline \newline JSON output: \\ \bottomrule
    
    \end{tabular}
    \caption{Prompt templates for CoLA.}
    \label{tab:CoLA-prompt-template}
\end{table*}

\begin{table*}[htb]
    \small
    \centering
    \begin{tabular}{p{5cm}p{10cm}}
    \toprule
    Task & Prompt template \\
    \midrule
    SingleClf & Classify which news category the following line of text belongs to among the following four categories: `Business,' `Sports,' `World,' and `Sci/Tech.'\newline \newline Text: \$text\newline News category: \\ \midrule
    
    BatchClf    & Classify which news category each of the \$num following lines of text belongs to among the following four categories: `Business,' `Sports,' `World,' and `Sci/Tech.'\newline \newline Texts, one per line:\newline \newline \$texts\newline \newline News categories for each of the \$num lines of text, one per line: \\ \midrule

    SelectOne & This is a news classification task in which each line of text belongs to one of four categories `Business,' `Sports,' `World,' and `Sci/Tech.'\newline \newline Go over the \$num lines of text below and list the index numbers of the lines that can be classified as \$category according to the following instructions: If none of the texts can be classified as \$category, write `None.' If all the texts can be classified as \$category, write `All.' Otherwise, provide the index numbers of the texts that can be classified as \$category.\newline \newline Output your responses in JSON format with the key `\$category'. A formatted example output is provided below.\newline \{`\$category': [None/All or index numbers of the texts that can be classified as \$category]\}\newline \newline Texts, one per line:\newline \newline \$texts\newline \newline JSON output: \\ \midrule

    SelectAll &  This is a news classification task in which each line of text belongs to one of four categories `Business,' `Sports,' `World,' and `Sci/Tech.'\newline \newline Go over the \$num lines of text below and list the index numbers of the lines that belong to each category according to the following instructions: If none of the texts can be classified as a particular category, write `None.' If all the texts can be classified as a particular category, write `All.' Otherwise, provide the index numbers of the texts that can be classified as the category.\newline\newline Output your responses in JSON format with the following keys: `business,' `sports,' `world,' and `sci/tech.' A formatted example output is provided below.\newline \{`business': [None/All or index numbers of texts in `business' category], `sports': [None/All or index numbers of texts in `sports' category], `world': [None/All or index numbers of texts in `world' category], `sci/tech': [None/All or index numbers of texts in sci/tech category]\}\newline \newline Texts, one per line:\newline \newline \$texts\newline \newline JSON output: \\ \bottomrule
    
    \end{tabular}
    \caption{Prompt templates for AGNews.}
    \label{tab:AGNews-prompt-template}
\end{table*}

\begin{table*}[htb]
    \small
    \centering
    \begin{tabular}{p{5cm}p{10cm}}
    \toprule
    Task & Prompt template \\
    \midrule
    SingleClf & Compare text A with text B and determine if text A is a paraphrase of text B. Respond with `Yes' if text A is a paraphrase, and `No' if it is not.\newline \newline \$text\newline Answer: \\ \midrule
    
    BatchClf    & Compare text A with text B for the following \$num text pairs and determine if text A is a paraphrase of text B line by line. Respond with `Yes' if text A is a paraphrase, and `No' if it is not. Provide your answers line by line.\newline \newline \$texts\newline Answers: \\ \midrule

    SelectOne & Go over the \$num text pairs below and list the index numbers of the text pairs where text A \$be a paraphrase of text B according to the following instructions: If none of the text pairs satisfy this condition, write `None.' If all the text pairs satisfy this condition, write `All.' Otherwise, provide the index numbers of the text pairs where text A \$be a paraphrase of text B.\newline \newline Output your responses in JSON format with the key `answer'. A formatted example output is provided below.\newline \{`answer': [None/All or index numbers of the text pairs where text A \$be a paraphrase of text B]\}\newline \newline Here are the text pairs:\newline \newline \$texts\newline JSON output: \\ \midrule

    SelectAll &  Go over the \$num text pairs below. First, list the index numbers of the text pairs that contain paraphrases. Then, list the index numbers of the text pairs that contain non-paraphrases. If none of the text pairs satisfy a condition, write `None.' If all the text pairs satisfy a condition, write `All.' Otherwise, provide the index numbers of the text pairs that satisfy each condition.\newline \newline Output your responses in JSON format with two keys: `yes' for paraphrases and `no' for non-paraphrases. A formatted example output is provided below.\newline \{`yes': [None/All or index numbers of text pairs that contain paraphrases], `no': [None/All or index numbers of text pairs that contain non-paraphrases]\}\newline \newline Here are the text pairs:\newline \newline \$texts\newline JSON output: \\ \bottomrule
    
    \end{tabular}
    \caption{Prompt templates for MRPC. For the text-pair classification task (MRPC, SNLI, WiC), the sequence of text pairs in the place of `\$texts' are indexed starting with `1' and each text pair is separated by two newlines (each text pair ends with a newline be design, followed by another newline before the next text pair).}
    \label{tab:MRPC-prompt-template}
\end{table*}

\begin{table*}[htb]
    \small
    \centering
    \begin{tabular}{p{5cm}p{10cm}}
    \toprule
    Task & Prompt template \\
    \midrule
    SingleClf & Given the following premise and hypothesis, determine the inference relation between them. Respond with `Entailment' if the hypothesis logically follows from the premise, `Contradiction' if they are in direct opposition, and `Neutral' if neither applies. \newline \newline \$text\newline Inference relation: \\ \midrule
    
    BatchClf    & Given the following \$num pairs of premises and hypotheses, determine the inference relation for each pair line by line. Respond with `Entailment' if the hypothesis entails the premise, and `Contradiction' if they contradict. If neither is the case, respond with `Neutral.' Provide your answers line by line.\newline \newline \$texts\newline Inference relations for the \$num text pairs provided above: \\ \midrule

    SelectOne & Go over the \$num text pairs below and list the index numbers of the text pairs where the inference relation between the premise and the hypothesis is \$relationship according to the following instructions: If none of the text pairs contain \$relationship inference relation, write `None.' If all text pairs contain \$relationship inference relation, write `All.' Otherwise, provide the index numbers of the text pairs where the inference relation between the premise and the hypothesis is \$relationship.\newline \newline Output your responses in JSON format with the key `\$relationship'. A formatted example output is provided below. \newline {`\$relationship': [None/All or index numbers of text pairs that contain \$relationship inference relation]} \newline \newline
    Here are the text pairs:\newline \newline \$texts\newline JSON output: \\ \midrule

    SelectAll &  Go over the \$num text pairs below. First, list the index numbers of the text pairs that contain entailment inference relation. Then, select all text pairs that contain contradiction inference relation. Finally, select all text pairs that contain neutral inference relation. If none of the text pairs satisfy a condition, write `None.' If all the text pairs belong satisfy a condition, write `All.' Otherwise, provide the index numbers of the text pairs that satisfy each condition.\newline \newline Output your responses in JSON format with three keys: `entailment', `contradiction', and `neutral'. A formatted example output is provided below.\newline \{`entailment': [None/All or index numbers of text pairs that contain entailment inference relation], `contradiction': [None/All or index numbers of text pairs that contain contradiction inference relation], `neutral': [None/All or index numbers of text pairs that contain neutral inference relation]\}\newline \newline Here are the text pairs:\newline \newline \$texts\newline JSON output: \\ \bottomrule
    
    \end{tabular}
    \caption{Prompt templates for SNLI.}
    \label{tab:SNLI-prompt-template}
\end{table*}

\begin{table*}[htb]
    \small
    \centering
    \begin{tabular}{p{5cm}p{10cm}}
    \toprule
    Task & Prompt template \\
    \midrule
    SingleClf & Analyze the usage of the given target word in the two subsequent contexts. The target word may appear in various grammatical forms in each context. Respond with `Yes' if it maintains the same meaning across both contexts, and `No' if it does not.\newline \newline \$text\newline Answer: \\ \midrule
    
    BatchClf    & Analyze the usage of the following \$num target words in the two contexts that immediately follow them. These target words may appear in different grammatical forms across the two subsequent contexts. Determine if each target word maintains the same meaning in the two subsequent contexts. Provide your answers line by line, indicating `Yes' if it does and `No' if it does not.\newline \newline \$texts\newline Answers: \\ \midrule

    SelectOne & Analyze the following \$num target words and determine the index numbers of the target words where the same meaning \$be maintained across the two contexts that immediately follow them. These target words may appear in different grammatical forms in each context. If none of the target words satisfy this condition, write `None.'. If all the target words satisfy this condition, write `All.' Otherwise, provide the index numbers.\newline \newline Output your responses in JSON format with the key `answer'. A formatted example output is provided below.\newline \{`answer': [None/All or index numbers of the target words where the same meaning \$be maintained in the two subsequent contexts]\}\newline \newline Here are the target words along with their contexts:\newline \newline \$texts\newline JSON output: \\ \midrule

    SelectAll &  Analyze the following \$num target words, which may appear in different grammatical forms in the two subsequent contexts. First, list the index numbers of target words that maintain the same meaning in the two subsequent contexts. Then, list the index numbers of target words that do not maintain the same meaning in the two subsequent contexts. If none of the target words satisfy a condition, write `None.' If all the target words satisfy a condition, write `All.' Otherwise, provide the index numbers of the target words that satisfy each condition.\newline \newline Output your responses in JSON format with two keys: `yes' for target words used with consistent meanings and `no' for those used with inconsistent meanings. A formatted example output is provided below.\newline \{`yes': [None/All or index numbers of target words used with consistent meanings], `no': [None/All or index numbers of target words used with inconsistent meanings]\}\newline \newline Here are the target words along with their contexts:\newline \newline \$texts\newline JSON output: \\ \bottomrule
    
    \end{tabular}
    \caption{Prompt templates for WiC.}
    \label{tab:WiC-prompt-template}
\end{table*}

\end{document}